%% file: main.tex
\documentclass{article}

\usepackage{microtype}
\usepackage{graphicx}
\usepackage{subcaption}
\usepackage{booktabs}

\usepackage[accepted]{icml2026}

\usepackage{amsmath}
\usepackage{amssymb}
\usepackage{mathtools}
\usepackage{amsthm}

\usepackage[capitalize,noabbrev]{cleveref}

\theoremstyle{plain}

\theoremstyle{definition}

\theoremstyle{remark}

\usepackage[disable,textsize=tiny]{todonotes}

\icmltitlerunning{
  Understanding Structured Health Data through
  Interaction-Aware Mixture-of-Experts
}

\begin{document}

\twocolumn[
  \icmltitle{
    Understanding Structured Health Data through \\
    Interaction-Aware Mixture-of-Experts
  }

  \begin{icmlauthorlist}
    \icmlauthor{Ji Hwan Park}{utaece}
    \icmlauthor{Ying Ding}{utasoi}
    \icmlauthor{Tianjin Guo}{iu}
  \end{icmlauthorlist}

  \icmlaffiliation{utaece}{
    Department of Electrical and Computer Engineering,
    The University of Texas at Austin,
    Austin, Texas, USA
  }

  \icmlaffiliation{utasoi}{
    School of Information,
    The University of Texas at Austin,
    Austin, Texas, USA
  }

  \icmlaffiliation{iu}{
    Kelley School of Business,
    Indiana University,
    Bloomington, Indiana, USA
  }

  % Replace this placeholder with the actual corresponding-author email.
  \icmlcorrespondingauthor{Ji Hwan Park}{your-email@utexas.edu}

  \icmlkeywords{
    Structured Health Data,
    Mixture-of-Experts,
    Clinical Prediction,
    Interpretability,
    Machine Learning
  }

  \vskip 0.3in
]

% Required by the ICML style.
\printAffiliationsAndNotice{}

\begin{abstract}
We study interaction-aware mixture-of-experts for post-stroke
rigidity prediction using multi-level views of structured health
records. Despite minimal performance gains, routing attribution
reveals systematic importance differences across views,
underscoring view construction as key to interpretability.
\end{abstract}

\input{main/introduction}
\input{main/related_works}
\input{main/method}
\input{main/results}
\input{main/conclusion}

%%%%%%%%%%%%%%%%%%%%%%%%%%%%%%%%
% ACKNOWLEDGEMENTS
%%%%%%%%%%%%%%%%%%%%%%%%%%%%%%%%

% Add acknowledgements for the camera-ready version if applicable.
%
% \section*{Acknowledgements}
% We thank ...

%%%%%%%%%%%%%%%%%%%%%%%%%%%%%%%%
% IMPACT STATEMENT
%%%%%%%%%%%%%%%%%%%%%%%%%%%%%%%%

\section*{Impact Statement}

This paper presents work whose goal is to advance the field of
machine learning for structured health data. Predictive models used
in healthcare may affect clinical decision-making and patient
outcomes. Their deployment therefore requires careful assessment of
generalizability, calibration, fairness, interpretability, privacy,
and potential distribution shifts. The methods studied in this work
are intended to support research and should not be used as a
substitute for professional clinical judgment without appropriate
external validation and oversight.

%%%%%%%%%%%%%%%%%%%%%%%%%%%%%%%%
% REFERENCES
%%%%%%%%%%%%%%%%%%%%%%%%%%%%%%%%

\bibliography{main/bib}
\bibliographystyle{icml2026}

%%%%%%%%%%%%%%%%%%%%%%%%%%%%%%%%
% APPENDIX
%%%%%%%%%%%%%%%%%%%%%%%%%%%%%%%%

% Uncomment if an appendix is needed.
%
% \newpage
% \appendix
% \onecolumn
%
% \section{Additional Results}
%
% Appendix material goes here.

\end{document}

%% file: main/introduction.tex
\section{Introduction}
\label{sec:intro}

Structured health data are a dominant substrate for clinical prediction,
spanning billing codes, diagnoses, procedures, vital signs, and other
irregular clinical measurements organized in tabular form
\citep{shickel2018deep,xu2025comprehensive}. Common modeling approaches
reflect this structure, as gradient-boosted decision trees and tabular
neural networks remain strong baselines on structured health-data tasks
\citep{chen2016xgboost,gorishniy2021ft,somepalli2021saint,popov2020node,wang2021dcn,shwartzziv2022tabular,mcelfresh2023neural}.
Yet most structured-data models either treat the record as a single
tabular input, leaving clinically meaningful interactions to be learned
implicitly, or attach naturally distinct modalities such as text or
imaging. It remains unclear whether transforming a single structured
record into multiple alternative representations, and modeling
interactions among them, can improve predictive performance while making
the prediction process easier to interpret.

To explore this question, we adopt the terminology of multi-view
learning, where a \textit{view} denotes an alternative representation of
the same underlying example~\citep{sun2013survey}. Multi-view learning
is designed to exploit complementary and shared information across views,
and has been applied broadly---for instance, to fuse imaging and clinical
text in medical diagnosis~\citep{wang2018tienet} or to combine different sensor
modalities in activity recognition~\citep{zhang2019hierarchical}. How to construct and
exploit multiple views of a \emph{single} structured health record,
however, remains underexplored.

For structured health data, such views can be constructed at multiple
levels. At the \textit{model level}, when the same structured input is
passed through different predictive models, each model's learned
representation constitutes a view, since different architectures encode
different inductive biases
\citep{gorishniy2021ft,popov2020node,shwartzziv2022tabular}. At the
\textit{data level}, the same structured input can be partitioned by
clinical semantics into administrative, procedural, diagnostic,
vital-sign, and code groups \citep{johnson2023mimiciv,ma2023ehr}. At the
\textit{representation level}, the record can be encoded through
different paradigms---graph-based patient--feature structure
\citep{brody2022gatv2,choi2020graphconvolutional,rocheteau2021predicting},
tabular representation learning, and text embeddings from
natural-language renderings of the record
\citep{hegselmann2023tabllm,lee2024meme,steinberg2021language}. Furthermore, While
simple concatenation of different views has the potential to improve predictive performance, it does
not distinguish view-specific signal from information shared across views
or emerging only through their combination, which can be critical for providing explanation in sensitive contexts such as healthcare.

To characterize these distinctions principally, we employ the Partial
Information Decomposition (PID) framework, which decomposes the
information that a set of views carries about a target into redundancy,
uniqueness, and synergy
\citep{williams2010nonnegative,bertschinger2014quantifying,liang2023quantifying}.
Specifically, we adopt the recently proposed I2MoE
framework~\citep{xin2025i2moe}, which routes inputs through specialized
experts and explicitly present view-specific and synergistic signals,
to study post-stroke rigidity prediction using a national inpatient
stroke cohort derived 129{,}401 hospital admissions. Using this setup, we examine whether
model-level, data-level, and representation-level views---all derived
from the same structured health record---provide useful interaction
structure for prediction and interpretation. 

Our results show that
introducing a multi-view approach yields minimal improvements in
predictive performance, consistent with views being alternative
decompositions of the same record rather than independent modalities.
Nevertheless, routing attribution reveals that model-level, data-level,
and representation-level view designs produce systematically different
allocations of importance across distinct views, and that these allocations are locally consistent across similar
patients. These findings suggest that view construction is a meaningful
design choice for interpretability even when predictive gains are modest,
and point toward routing-based attribution as a practical lens for
building explanations from structured health records.

%% file: main/related_works.tex
\section{Related Work}
\label{sec:related}

\paragraph{Structured tabular health modeling.}
Structured health data are central to clinical prediction because they
capture diagnoses, procedures, utilization, and longitudinal
measurements at scale~\citep{shickel2018deep,xu2025comprehensive}.
These data are commonly modeled as a single tabular representation.
Gradient-boosted trees, including XGBoost, remain strong baselines for
tabular prediction~\citep{chen2016xgboost}, while tabular neural
architectures such as SAINT, NODE, and DCN-V2 provide competitive
alternatives for structured inputs
\citep{somepalli2021saint,popov2020node,wang2021dcn}. These methods, however, largely preserve a
single-view treatment of the record and abstract away from the distinct clinical concepts each measure represent. We ask whether a single structured
health record can instead be expressed through multiple view
definitions, and whether those views make interaction-aware prediction
and interpretation more informative.

\vspace{-0.5em}
\paragraph{Interaction modeling and interpretability.}
Clinical prediction from structured records often depends on how
variables act together rather than in isolation. Interpretable additive
models with pairwise terms have shown that selected feature interactions
can support clinical risk prediction while remaining inspectable
\citep{caruana2015intelligible}, while neural interaction detection and
post-hoc attribution methods explain dependencies in trained predictors
\citep{tsang2018detecting,lundberg2017unified}. These approaches
primarily operate at the level of variables or variable pairs. In
contrast, view-based modeling organizes a structured record into
higher-level sources before fusion, shifting the question to whether
each view contributes distinct, duplicated, or jointly useful signal.
Partial Information Decomposition formalizes these cases as unique,
redundant, and synergistic information
\citep{williams2010nonnegative,bertschinger2014quantifying,liang2023quantifying}.
I2MoE implements this idea in a predictive MoE through unique, synergy,
and redundancy experts~\citep{xin2025i2moe}. Our work uses this
framework to evaluate whether view-level interaction modeling is useful when
all views are constructed from the same structured health record.

%% file: main/method.tex
\section{Methods}
\label{sec:method}

\begin{table*}[t]
  \caption{Summary of predictive performance. $^*$ denotes a
    statistically significant improvement over XGBoost across random
    seeds.}
  \label{tab:main}
  % \vskip 0.1in
  \begin{center}\begin{small}
  \begin{tabular}{lllccc}
    \toprule
    \textbf{Model} & \textbf{Type} & \textbf{Setting} & \textbf{AUROC} & \textbf{AUPRC} & \textbf{F1} \\
    \midrule
    XGBoost        & Machine Learning & --              & $0.7627$ & $0.7786$ & $0.7182$ \\
    GATv2          & Graph-Based      & --              & $0.7512$ & $0.7578$ & $0.7235$ \\
    SAINT          & Tabular DL       & --              & $0.7625$ & $0.7789$ & $0.7170$ \\
    NODE           & Tabular DL       & --              & $0.7634$ & $0.7799$ & $0.7160$ \\
    DCN-V2         & Tabular DL       & --              & $0.7626$ & $0.7789$ & $0.7154$ \\
    \midrule
   $\text{I}^2\text{MoE}$ (Model-Level)  & Mixture of Experts & Full          & $0.7714^*$ & $0.7888^*$ & $0.7213$ \\
                                           &                        & No Synergy    & $0.7713^*$ & $0.7887^*$ & $0.7223$ \\
                                           &                        & No Redundancy & $0.7713^*$ & $0.7888^*$ & $0.7226$ \\
                                           &                        & No S \& R     & $0.7713^*$ & $0.7886^*$ & $0.7171$ \\
    \midrule
    $\text{I}^2\text{MoE}$ (Data-Level)    & Mixture of Experts & Full          & $0.7621$ & $0.7788$ & $0.7182$ \\
                                           &                        & No Synergy    & $0.7621$ & $0.7787$ & $0.7185$ \\
                                           &                        & No Redundancy & $0.7621$ & $0.7788$ & $0.7186$ \\
                                           &                        & No S \& R     & $0.7622$ & $0.7786$ & $0.7178$ \\
    \midrule
    $\text{I}^2\text{MoE}$ (Representation-Level) & Mixture of Experts & Full          & $0.7600$ & $0.7781$ & $0.7313^*$ \\
                                           &                        & No Synergy    & $0.7600$ & $0.7781$ & $0.7317^*$ \\
                                           &                        & No Redundancy & $0.7601$ & $0.7781$ & $0.7307^*$ \\
                                           &                        & No S \& R     & $0.7601$ & $0.7781$ & $0.7316^*$ \\
    \bottomrule
  \end{tabular}
  \end{small}\end{center}
  \vskip -0.15in
\end{table*}

\subsection{Task Definition}
\label{sec:task}

To study how interaction-aware modeling behaves when a single structured
record is expressed through alternative views, we use post-stroke
rigidity prediction as a binary classification task. The cohort is
derived from adult HCUP/NIS stroke hospitalizations from 2016--2020, consisting of 129{,}401 admissions.
HCUP/NIS is an all-payer database of U.S. hospital inpatient stays
derived from hospital billing data and includes clinical and
resource-use information typically available from discharge abstracts.
We define rigidity using a clinician-curated set of 55 ICD-10-CM codes. Selected predictor variables span administrative factors,
procedures, diagnoses, clinical signs, and grouped ICD indicators.

\subsection{Three View Definitions}
\label{sec:three_levels}

For structured health data, the view structure is often a modeling
choice rather than a fixed property of the raw record. The structured
nature of these records allows alternative views to be defined at
different stages of the modeling pipeline: before modeling by grouping
variables, during modeling through architecture-specific encoders, and
after upstream encoding by fusing learned representations. We use these
three stages to define data-level, model-level, and
representation-level views while keeping the I2MoE formulation fixed.

\subsubsection{Model-Level Views}
\label{sec:model_level}

The model-level formulation treats representations from tabular
prediction backbones as views. We encode the full processed record with
SAINT, NODE, and DCN-V2
\citep{somepalli2021saint,popov2020node,wang2021dcn}, three models
designed for structured tabular inputs with different inductive biases:
attention-based feature modeling, tree-inspired representation learning,
and explicit feature crossing. Their hidden representations are passed
to separate unique experts, evaluating whether tabular-specialized models
produce distinct predictive signals from the same record.

\subsubsection{Data-Level Views}
\label{sec:data_level}

The data-level formulation defines views by semantic feature partition.
The record is split into administrative, procedure, diagnosis, clinical
sign, and ICD-indicator groups. Each group is encoded by a separate
multilayer perceptron unique expert. This examines whether interaction
structure emerges when views correspond to clinically meaningful
subdomains of the same structured record.

\subsubsection{Representation-Level Views}
\label{sec:representation_level}

The representation-level formulation defines views after upstream
encoding. For each patient, we use a frozen GATv2 encoder
\citep{brody2022gatv2} to obtain a graph representation, a frozen NODE
encoder~\citep{popov2020node} to obtain a tabular representation, and a
frozen Qwen3-Embedding encoder~\citep{zhang2025qwen3embedding} applied
to a structured-text rendering of the record to obtain a text
representation. These embeddings are then fused by I2MoE. This evaluates
whether interaction-aware fusion is more useful after view-specific
representation learning.

\subsection{I2MoE Formulation and Objective}
\label{sec:imoe}

Let $\{\mathbf{h}_m\}_{m=1}^{M}$ denote the view-specific
representations, with $M{=}3$ for model-level and representation-level
views and $M{=}5$ for data-level views. Following I2MoE
\citep{xin2025i2moe}, we use one unique expert per view, one synergy
expert, one redundancy expert, and a reweighting network $g$:
\vspace{-0.7em}
\begin{equation}
  \hat{y} \;=\; \sigma\!\left(
    \sum_{m=1}^{M} w_m^u f_m^u(\mathbf{h}_m)
    + w^s f^s(\mathbf{h})
    + w^r f^r(\mathbf{h})
  \right),
  \label{eq:imoe}
\end{equation}
where
$[w_1^u,\dots,w_M^u,w^s,w^r] = g([\mathbf{h}_1;\dots;\mathbf{h}_M])$.
The synergy expert is intended to capture signal that emerges only when
views are considered jointly, whereas the redundancy expert captures
signal shared across views.

We optimize the task loss and the weakly supervised I2MoE interaction
loss, which encourages unique, synergy, and redundancy experts to
capture view-specific, combined, and shared information:
\vspace{-0.75em}
\begin{equation}
  \mathcal{L}_{\mathrm{total}}
  \;=\;
  \mathcal{L}_{\mathrm{task}}
  +
  \lambda_{\mathrm{int}} \mathcal{L}_{\mathrm{int}},
  \label{eq:loss}
\end{equation}
We refer readers to \citet{xin2025i2moe} for the full construction of
$\mathcal{L}_{\mathrm{int}}$.

\subsection{Experimental Protocol}
\label{sec:protocol}

For each view definition, we train a full I2MoE model with unique,
synergy, and redundancy experts, together with three ablations that
remove the synergy expert, the redundancy expert, or both. We evaluate
predictive performance using AUROC, AUPRC, and F1, with all predictive
results averaged over 30 random seeds. We use expert-routing weights for interpretation at two levels. Global
interpretation summarizes how each view definition allocates predictive
mass across unique, synergy, and redundancy experts at the cohort level,
estimated by averaging expert-routing weights over the test cohort and
across retrainings. Local interpretation asks whether patients with
similar learned representations receive similar expert-weight
allocations. For this, we use the representation-level model and compare
expert-routing distances between nearest-neighbor patients in the
learned representation space and randomly matched patients.

%% file: main/results.tex
\section{Experiments and Results}
\label{sec:results}

\subsection{Predictive Performance}
\label{sec:main_results}

The three view definitions are competitive with strong graph and tabular
baselines, but their advantages are metric-dependent (\cref{tab:main}).
Compared with XGBoost, the model-level view shows a statistically
significant increase in AUROC and AUPRC, while the representation-level
view shows a statistically significant increase in F1 but lower AUROC.
The data-level view is not significantly different from XGBoost on the
main metrics. This is the central empirical pattern of our study. A
single structured health record can be re-expressed as model-level,
data-level, or representation-level views, and these choices change
prediction behavior. However, the performance gains are limited because
the views are alternative decompositions of the same underlying record
rather than independent modalities.

\subsection{Interaction Ablation}
\label{sec:ablation}

The interaction ablations in \cref{tab:main} show no statistically
significant degradation after removing synergy, redundancy, or both.
This suggests that the interaction experts are not the main source of
the observed predictive performance. Instead, most predictive signal
appears recoverable from the unique experts and the reweighting network.
In this setting, synergy and redundancy experts are more diagnostic than
performance-improving: they expose how the model allocates mass to
view-specific, shared, and combined signals, but do not by themselves
produce a clear predictive gain. This suggests that, when views are
constructed from the same structured record rather than separate data
sources, extracting additional predictive signal may require stronger
interaction modeling or more distinctive view construction.

\begin{figure}[t]
  \vspace{-0.6em}
  \centering
  \includegraphics[width=0.98\columnwidth]{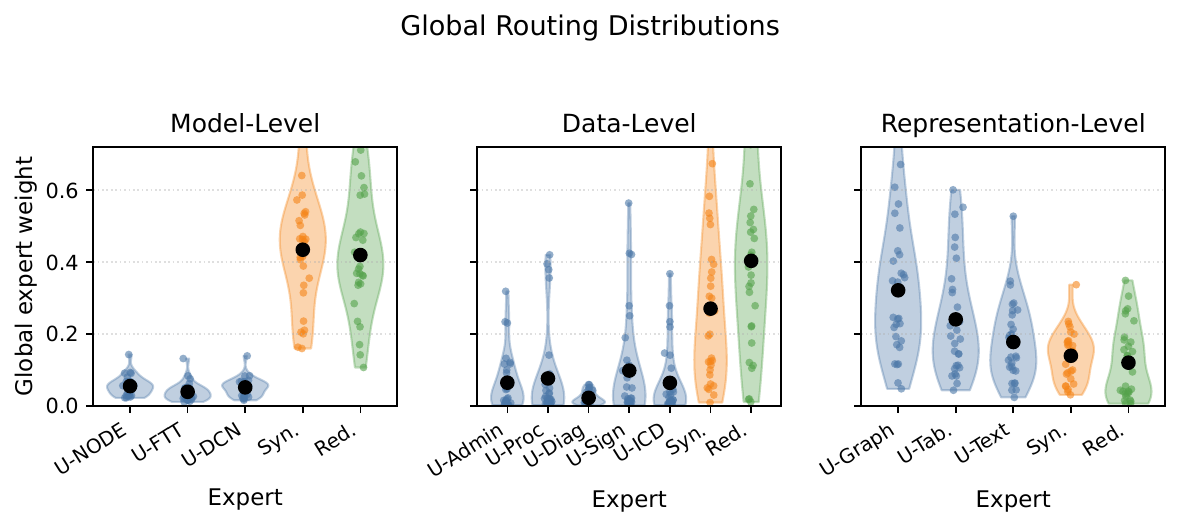}
  \vspace{-0.6em}
  \caption{Cohort-level expert routing across retrainings.}
  \label{fig:global_routing}
  \vspace{-2.5em}
\end{figure}

\subsection{Global Interpretation}
\label{sec:stability}

I2MoE provides routing-based interpretation through a reweighting network
that assigns an expert weight to each unique, synergy, and redundancy
expert for each prediction. Averaging these weights over the test cohort
summarizes how the model allocates prediction across the view-specific,
shared, and jointly useful signals defined in \cref{sec:imoe}
(\cref{fig:global_routing}). Model-level views place high mass on both
synergy and redundancy experts. This is consistent with the fact that
the views are different tabular backbones trained on the same input,
whose learned representations can overlap substantially while still
producing complementary decision patterns. Data-level views also route
substantial mass through synergy and redundancy experts, reflecting the
dependence among clinical groups such as diagnoses, procedures, signs,
and administrative factors. These groups are clinically distinct but not
statistically independent, so shared and combined signal are expected.
In contrast, representation-level views allocate more mass to unique
experts, suggesting that graph, tabular, and text-derived embeddings
preserve more view-specific signal. Overall, global expert weights show
that view definition changes how the model distributes prediction across
view-specific and interaction experts. We interpret these weights as
model allocation patterns rather than direct clinical evidence of
synergy or redundancy, since all views are derived from the same
structured record.

\subsection{Local Interpretation}

\begin{figure}[t]
  \vspace{-0.6em}
  \centering
  \includegraphics[width=0.98\columnwidth]{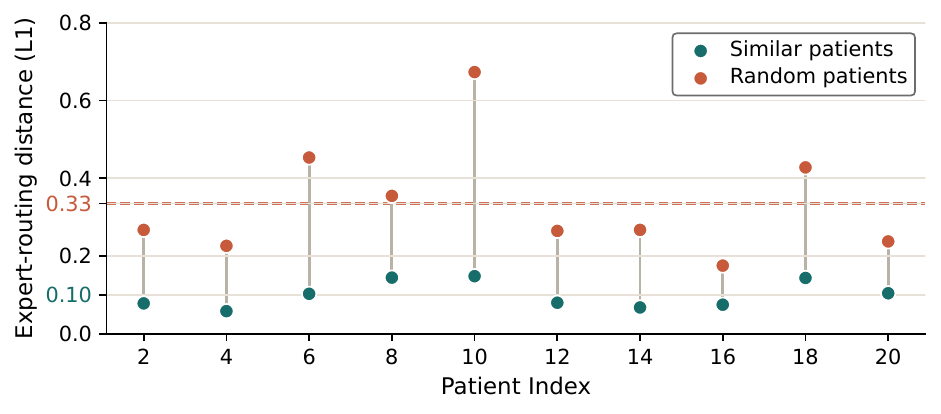}
  \vspace{-0.6em}
  \caption{Patient-level routing consistency. 
  %Similar patients receive more similar expert weights than random patients.
  }
  \label{fig:local_routing}
  \vspace{-2.3em}
\end{figure}

We further study how the model assigns expert weights for individual
patients through the I2MoE reweighting network. This analysis is
performed for the representation-level model, where each patient has a
learned graph-tabular-text representation. To study patient-neighborhood
behavior, we select 10 evaluation patients and use each one to define a
local neighborhood in this representation space. For each selected
patient, we compare the expert-weight distributions of its nearest
neighbors with those of randomly matched patients
(\cref{fig:local_routing}). Nearby patients have consistently smaller
expert-weight distance than random controls, suggesting that
patient-level expert weights vary in a locally consistent way
rather than changing arbitrarily across similar cases. Such local consistency is an encouraging property that enables future work on utilizing expert routing information to provide meaningful patient-level explanations.

%% file: main/conclusion.tex
\section{Conclusion}
\label{sec:conclusion}

We examined interaction-aware mixture-of-experts modeling for
post-stroke rigidity prediction when one structured health record is
expressed through model-level, data-level, and representation-level
views. Multi-view modeling yields metric-dependent but limited predictive gains, 
reflecting that the views are alternative representations of the same structured health record 
rather than independent modalities. Furthermore, removing interaction experts does not
significantly degrade performance---suggesting that explicit synergy and
redundancy modeling is more useful for diagnosing how views are used
than for improving prediction. Routing analyses reveal that view
construction choices produce systematically different expert allocations
at the cohort level, and that these allocations are locally consistent
across similar patients. Together, these findings position routing-based
attribution as a promising foundation for offering accurate prediction with 
patient-level explanation in healthcare context based on structured data. 
Future work in this direction can aim to connect expert-level routing
to finer-grained clinical concepts, and developing view constructions
that extract more distinctive signal from the same underlying record.